\documentclass[sigconf,nonacm,anonymous,9pt]{IEEEtran}
\usepackage{subcaption}
\usepackage{booktabs}
\usepackage{multirow} 
\usepackage{tikz}
\usepackage{colortbl}
\usepackage{cite}
\usepackage{amsmath,amssymb,amsfonts}
\usepackage{algorithmic}
\usepackage{graphicx}
\usepackage{textcomp}
\usepackage{xcolor}
\usepackage{cleveref}
\usepackage{helvet}
\usepackage{siunitx}
\usepackage{spverbatim}

\definecolor{lightgreen}{HTML}{73B405}

\definecolor{myorange}{RGB}{244,106,18} 
\definecolor{myblue}{RGB}{0,111,190}    
\definecolor{mygreen}{RGB}{0,127,128}   
\definecolor{myred}{RGB}{228,46,36}     
\definecolor{myyellow}{RGB}{198,148,34} 
\definecolor{mydark}{RGB}{114,44,114}   
\definecolor{mymiddle}{RGB}{144,44,144} 
\definecolor{mylight}{RGB}{167,44,167}  
\usepackage[compact]{titlesec}
\usepackage[flushleft]{threeparttable}
\titlespacing*{\section}
{1pt}{1ex}{.5ex}
\titlespacing*{\subsection}
{1pt}{.1ex}{.5ex}
\titlespacing{\subsubsection}{.5pt}{0ex}{0ex}

\usepackage{amsmath,amsfonts,bm}

\def\eqref#1{equation~\ref{#1}}

\def\1{\bm{1}}

\def\vh{{\bm{h}}}

\def\vq{{\bm{q}}}

\DeclareMathAlphabet{\mathsfit}{\encodingdefault}{\sfdefault}{m}{sl}
\SetMathAlphabet{\mathsfit}{bold}{\encodingdefault}{\sfdefault}{bx}{n}

\def\gE{{\mathcal{E}}}

\def\gG{{\mathcal{G}}}

\def\gV{{\mathcal{V}}}

\def\chi{{\mathcal{X}}}

\newcommand{\R}{\mathbb{R}}

\graphicspath{{./figs/}}

\definecolor{mypurple}{HTML}{b05f8c}
\definecolor{mypink}  {HTML}{f58297}
\definecolor{myyellow}{HTML}{f7d35c}
\definecolor{lightpink}{HTML}{FFBFBF}
\definecolor{lightyellow}{HTML}{FFEF93}
\definecolor{cmured}{HTML}{AC232E}
\definecolor{lightpurple}{HTML}{E4C1EA}
\definecolor{gptgreen}{HTML}{6DA697}
\definecolor{lightblue}{HTML}{629CD3}
\definecolor{myorange}{HTML}{F46A12}
\newcommand\greencircle[1]{%
  \tikz[baseline=(X.base)] 
    \node (X) [draw=none, shape=circle, inner sep=0, fill=gptgreen, text=white] {\sffamily\strut #1};%
}
\newcommand\blackcircle[1]{%
  \tikz[baseline=(X.base)] 
    \node (X) [draw=none, shape=circle, inner sep=0, fill=black, text=white] {\sffamily\strut #1};%
}
\begin{document}

\title{$\mathsf{BRIDGES}$: \underline{Brid}ging \underline{G}raph Modality and \\ Large Language Models within \underline{E}DA Task\underline{s}}


\author{%
  \IEEEauthorblockN{Wei Li, Yang Zou, Christopher Ellis, Ruben Purdy, Shawn Blanton, Jos\'e M. F. Moura \\}
  \IEEEauthorblockA{Department of Electrical and Computer Engineering, Carnegie Mellon University}
}
\maketitle
\begin{abstract}
\label{sec:abstract}
While many EDA tasks already involve graph-based data, 
existing LLMs in EDA primarily either represent graphs as sequential text, or simply ignore graph-structured data that might be beneficial like dataflow graphs of RTL code.
Recent studies have found that LLM performance suffers when graphs are represented as sequential text, 
and using additional graph information significantly boosts performance. 
To address these challenges, we introduce $\mathsf{BRIDGES}$, a framework designed to incorporate graph modality into LLMs for EDA tasks.
$\mathsf{BRIDGES}$ integrates an automated data generation workflow, a solution that combines graph modality with LLM, and a comprehensive evaluation suite. 
First, we establish an LLM-driven workflow to generate RTL and netlist-level data, converting them into dataflow and netlist graphs with function descriptions. 
This workflow yields a large-scale dataset comprising over 500,000 graph instances and more than 1.5 billion tokens. 
Second, we propose a lightweight cross-modal projector that encodes graph representations into text-compatible prompts, enabling LLMs to effectively utilize graph data without architectural modifications.
Experimental results demonstrate 2x to 10x improvements across multiple tasks compared to text-only baselines, 
including design retrieval, type prediction, function description, and power/area estimation,
with negligible computational overhead ($<$1\% model weights increase and $<$30\% additional runtime overhead).
Even without additional LLM fine-tuning, our results outperform text-only and graph-only by a large margin.
We plan to release $\mathsf{BRIDGES}$, including the dataset, models, and training flow. 
\end{abstract}

\section{Introduction}
\label{sec:intro}

Recent advancements of large language models (LLMs) 
demonstrate remarkable scalability, adaptability to varied scenarios, and enhanced reasoning capabilities \cite{brown2020language, bubeck2023sparks, wei2022chain}. 
These strengths align well with the demands of modern Electronic Design Automation (EDA), creating an exciting opportunity
 to revolutionize the EDA workflow.
 Consequently, researchers have begun exploring LLM applications in EDA, 
 including tasks like Register-Transfer Level (RTL) code generation \cite{liu2024rtlcoder, pei2024betterv, chang2024data, chang2024natural}, 
 RTL debugging \cite{xu2024meic, yao2024rtlrewriter}, API recommendations \cite{wu2024chateda}, and document-based question-answering \cite{pu2024customized}.

However, a fundamental challenge exists: while LLMs handle text sequences well, many EDA tasks involve graph-based data (e.g., logic netlists) 
or benefit from graph representations such as dataflow graphs for RTL \cite{guo2020graphcodebert} and layout graph for layout decomposition\cite{li2022rethinking}.
Representing graphs using natural language (\textit{e.g.}, the netlist Verilog (.v) file) is called \textit{Graph2Text} \cite{wang2024can}.
Recent studies have found that LLM performance suffers when graph structures are presented as sequential text \cite{chai2023graphllm, liu2024lost}.
There are two main issues: 
First, LLMs struggle to fully interpret and learn from essential graph properties, like structure and functionality, when they are encoded as linear sequences.
Second, representing complex graphs as text often leads to overly lengthy contexts, making it challenging for LLMs to identify key information for accurate reasoning.
These limitations are especially pronounced in EDA, where graph scale is considerably larger than in other domains.
Further, when graph structures contain implicit but critical knowledge, 
adding this information through graph representations significantly boosts model performance. 
For instance, GraphCodeBERT \cite{guo2020graphcodebert} incorporates code dataflow graphs into the LLM pre-training process, achieving far superior results compared to text-only LLMs.

\begin{table}[]
    \centering
    \resizebox{.45\textwidth}{!}{\begin{tabular}{c|c|c|c}
    \toprule
        Circuit & File length & GPT-4o & Claude 3.5 Sonnet \\
    \midrule

        16-bit adder &4,119 & {\color{gptgreen}\checkmark} & {\color{gptgreen}\checkmark} \\
        16-bit comparator & 1,372& {\color{gptgreen}\checkmark} & {\color{gptgreen}\checkmark} \\
        8-bit divider & 27,137 & {\color{cmured} 8-bit comparator} & {\color{cmured} 8-bit comparator} \\
        16-bit divider &108,589 & {\color{cmured}16-bit multiplier} & {\color{cmured}16-bit multiplier} \\
        8-bit multiplier & 26,502& {\color{gptgreen}\checkmark} & {\color{gptgreen}\checkmark} \\
        16-bit multiplier &112,636 & \color{gptgreen}\checkmark & {\color{cmured} 8-bit comparator} \\ 
    \bottomrule
    \end{tabular}}
    \caption{Bit and type prediction using netlist file.}
    \label{tab:toy}
    \vspace{-4mm}
\end{table}

We argue that text-only approaches pose even greater limitations in EDA.
To illustrate this, we conduct a preliminary experiment (Table \ref{tab:toy}), in which two state-of-the-art commercial LLMs, GPT-4o and Claude 3.5 Sonnet, 
are tasked with predicting bit-widths and arithmetic types from netlist files.\footnote{See Appendix B for prompts used.
}
The results reveal their underperformance, particularly with longer input contexts.
 However, it is not easy to introduce graph modality into LLMs for various EDA tasks.
 We attribute these challenges to
 1) a lack of large-scale EDA-specific graph datasets, 
 2) limited exploration of graph integration in LLMs for large graphs typical of EDA, 
 and 3) absence of appropriate methods to evaluate graph-based LLM performance.

To address the limitations outlined above, this paper introduces $\mathsf{BRIDGES}$ (Bridging Graph Modality and
Large Language Models within EDA Tasks), a comprehensive framework that integrates graph modalities into LLMs to enhance their application in EDA tasks, covering automated data generation, model design, training, and evaluation. 
First, $\mathsf{BRIDGES}$ includes an automated graph-format data generation flow using LLMs and EDA tools. 
This workflow builds on the RTL code generation methods introduced in RTLCoder \cite{liu2024rtlcoder} and MG-Verilog \cite{zhang2024mg},
which generates RTL code and function description automatically. 
Specifically, the generated RTL code is transformed into a series of dataflow graphs and synthesized into netlist graphs using EDA tools;
the description is used to determine circuit type.
Together, these elements form an enriched multi-modality data instance—comprising the RTL code, dataflow graph, netlist graphs, function description, PPA metrics, and circuit type—for each RTL example. 
Through this process, $\mathsf{BRIDGES}$ has generated a large-scale dataset with over 500,000 graphs, containing more than 1.5 billion tokens.
To integrate text and graph modalities, we introduce a light-weight cross-modal projector that maps graph representations into soft prompts in the text space, 
thereby enabling LLMs to effectively process graph, especially the large-scale graphs typical of EDA. 
Finally, we conduct extensive experiments to validate the ability of $\mathsf{BRIDGES}$ to understand and interpret VLSI designs across tasks, including design-function retrieval, circuit type prediction, function description generation, and area/power estimation.
We summarize our contributions:

\begin{itemize}
    \item We introduce a novel LLM-driven workflow for generating and integrating RTL and netlist-level data, along with their graph-based representations and relevant attributes, such as function descriptions.
    \item Leveraging this workflow, we construct a large-scale dataset comprising over 37,000 data instances and 500,000 graphs, encompassing more than 1.5 billion tokens.
    \item Together with the data generation flow, we introduce \textbf{$\mathsf{BRIDGES}$}, the first method in VLSI domain to integrate graph modality, enhancing LLM performance across diverse EDA tasks.
    \item We conduct extensive experiments on four tasks and observe a several-fold performance improvement, demonstrating the importance of graph modality and effectiveness of $\mathsf{BRIDGES}$.
    \item $\mathsf{BRIDGES}$ will be fully open-sourced once published, including the data generation process, the entire generated dataset, training algorithms, the graph-supported LLM with a cross-modal projector, and the fine-tuned model.
\end{itemize}
\section{Related Work}
\label{sec:related}
\subsection{LLMs in EDA}
Most existing LLM work in EDA primarily leverages the text modality.
These approaches can be categorized based on how they use LLMs,
including but not limited to prompt engineering, fine-tuning, and retrieval-based techniques.

Prompt engineering involves evaluating different prompts to enhance LLM performance, sometimes with a systematic approach for using varied prompts across stages. 
For example, Thakur et al. \cite{thakur2023autochip} proposes an automated HDL generation framework that iteratively updates prompts based on feedback from a code evaluator. 
Similarly, ChipGPT \cite{chang2023chipgpt} uses ChatGPT to assess the capacity of an LLM to generate hardware logic from natural-language specifications, employing a prompt manager for portability.
In ChipChat \cite{blocklove2023chip}, the authors use ChatGPT-4 interactively to design an 8-bit accumulator-based microprocessor architecture, 
resulting in the first fully AI-generated HDL for tapeout. 
However, prompt engineering often falls short in tasks that require a high-quality result.

Beyond prompt engineering, some studies enhance LLMs for downstream tasks through data collection and fine-tuning, particularly in RTL code generation \cite{chang2024data,pei2024betterv,liu2024rtlcoder,zhang2024mg, liu2023verilogeval} and RTL debugging \cite{xu2024meic}. 
RTLCoder \cite{liu2024rtlcoder} generates over 27,000 RTL-instruction pairs in an automated flow, while MG-Verilog \cite{zhang2024mg} presents an open-source dataset of RTL code and descriptions at multiple granularity.
They use the collected data to fine-tune LLMs and is able to produce comparable or even better RTL code than commercial GPT-4 in VerilogEval benchmark \cite{liu2023verilogeval}. 
BetterV \cite{pei2024betterv} further improves the quality of generated RTL code through instruction-tuning and data augmentation.
While promising, these models remain less efficient and effective than human experts.   
We argue that the main reason is that text modality alone does not sufficiently capture the subtle properties of large-scale VLSI designs, that includes, for example, the critical path and the repeat among some local sub-circuits.
We believe enhancing the understanding of dataflow graphs hidden in an ocean of existing RTL code, the LLM agent could significantly improve its RTL-related performance, e.g., RTL code generation and debugging.

Retrieval-based techniques \cite{yao2024rtlrewriter,xiong2024hlspilot,pu2024customized,yin2024ado} are also widely deployed to boost the effectiveness of LLMs within EDA.
Pu et al. \cite{pu2024customized} creates a database from two textbooks and the OpenROAD project documentation, enabling more accurate EDA tool usage responses.
HLSPilot \cite{xiong2024hlspilot} applies a retrieval-based method using Xilinx manuals for HLS optimization, 
facilitating optimized HLS code generation via LLMs. 
These approaches demonstrate strong potential, especially with vast unstructured industrial data. 
However, text is not a sufficient representation for long-context, highly structured, graph-like data. 
Our experiments demonstrate that graph representations can substantially boost retrieval performance.

Some efforts have been made to introduce a visual modality into LLMs for EDA.
For example, Chang et al. \cite{chang2024natural} points out the deficiency of the text-only modality in the RTL code generation task
and therefore uses the design diagram to augment the description, which improves the quality of generate RTL code.
Yao et al. \cite{yao2024rtlrewriter} instead uses the visual diagram of the critical path to help optimize the RTL code.
Very recently, CircuitFusion \cite{fang2025circuitfusion} is proposed to fuse graph modality to embed the design, but the application and impact of graph modality to LLMs are still under-explored.

\subsection{Graph modality in LLM}

Although graph modalities are rarely used in EDA, they have been applied in other fields \cite{cao2023instructmol,liu2023molca,tang2024graphgpt,chai2023graphllm}.
MolCA \cite{liu2023molca} represents molecules as graphs for property prediction and description tasks, while InstructMol \cite{cao2023instructmol} combines graph-based representations with instruction-tuning to assist in molecular analysis.
In other contexts, GraphGPT \cite{tang2024graphgpt} and GraphLLM \cite{chai2023graphllm} show that integrating graph knowledge improves graph reasoning. 
GraphCodeBERT \cite{guo2020graphcodebert} uses dataflow graphs during pre-training, achieving notable gains over text-only approaches. 
These results suggest that incorporating graph modality could enhance LLM performance in EDA applications.
However, the large-scale, complex graphs typical of EDA, and the lack of large-scale graph datasets in EDA, pose unique challenges that require specialized solutions.
\section{Dataset Generation}
\begin{figure}
    \centering
    \includegraphics[width=0.52\textwidth]{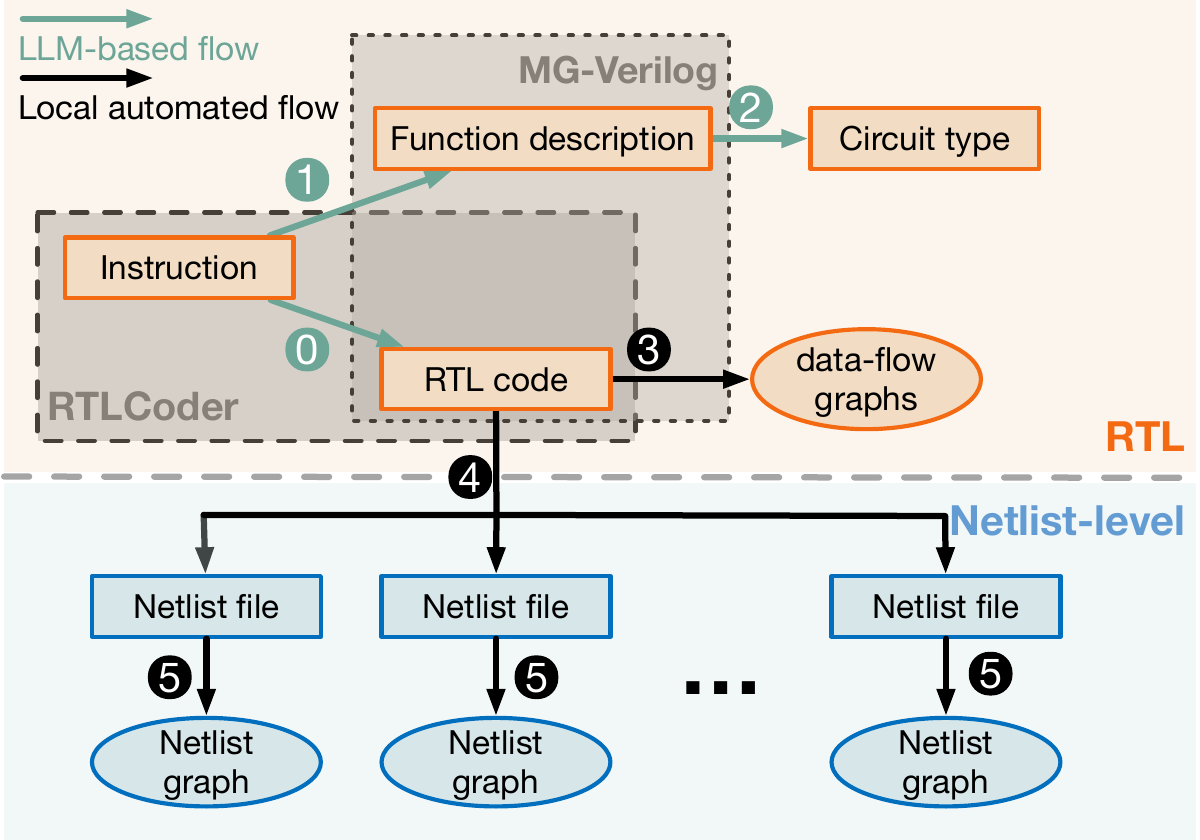}
    \caption{Automatic dataset generation workflow based on extensions of RTLCoder and MG-Verilog.}
    \label{fig:data_generation}
    \vspace{-4mm}
\end{figure}

$\mathsf{BRIDGES}$ introduces an automated workflow for generating RTL and netlist data in both text and graph modalities. 
Specifically, \Cref{fig:data_generation} illustrates the workflow that integrates EDA tools and LLMs to create extensive datasets for training and evaluating LLMs augmented with graph modality.
Each data instance includes, RTL code, its function description and circuit-type label.
Additionally, it encompasses dataflow graphs of RTL, 27 different netlist implementations, corresponding netlist graphs and PPA information. 
An example of each element in the data instance is shown in \Cref{fig:data_example} (left).
Data generation process is detailed in Appendix A.

\begin{figure*}
    \centering
    \includegraphics[width=1.05\textwidth]{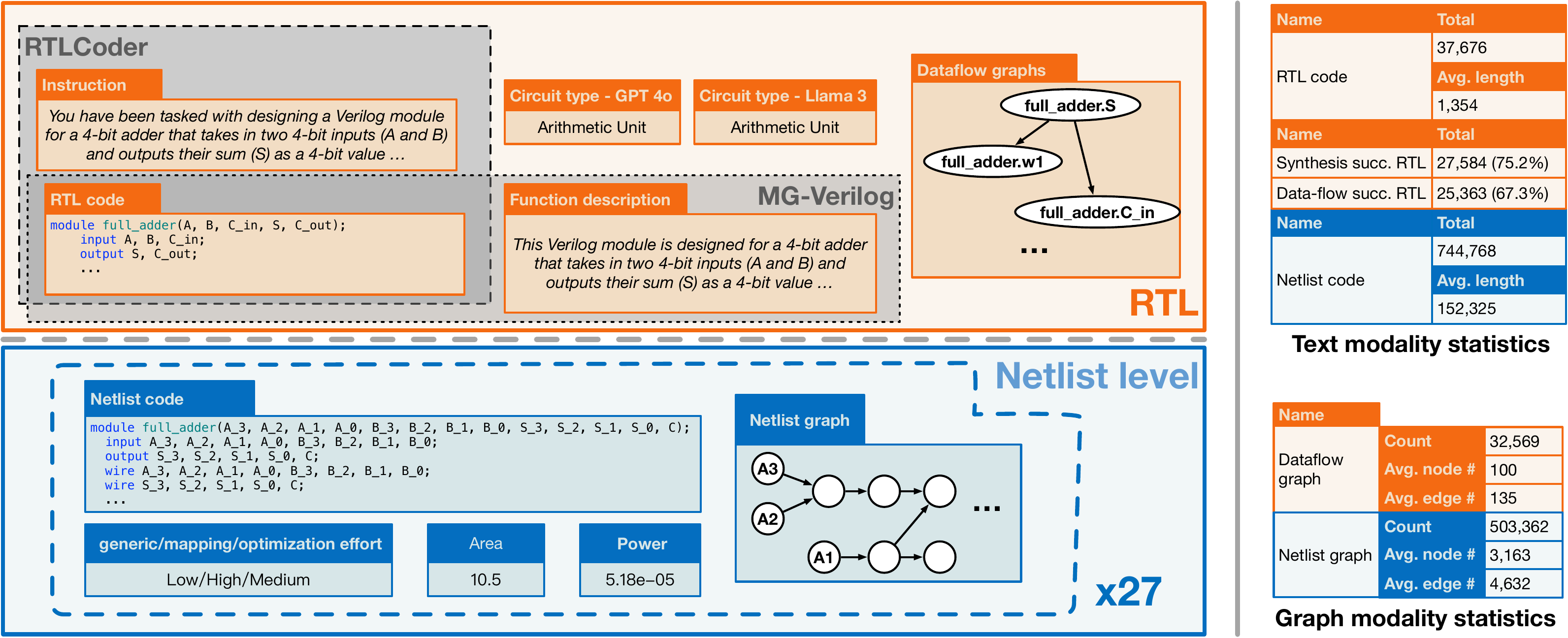}
    \caption{\textbf{Left}: Example of an enriched multi-modal data instance in $\mathsf{BRIDGES}$. \textbf{Right}: Statistics of the dataset.}
    \label{fig:data_example}
    \vspace{-4mm}
\end{figure*}

\begin{figure}
    \centering
    \includegraphics[width=0.5\textwidth]{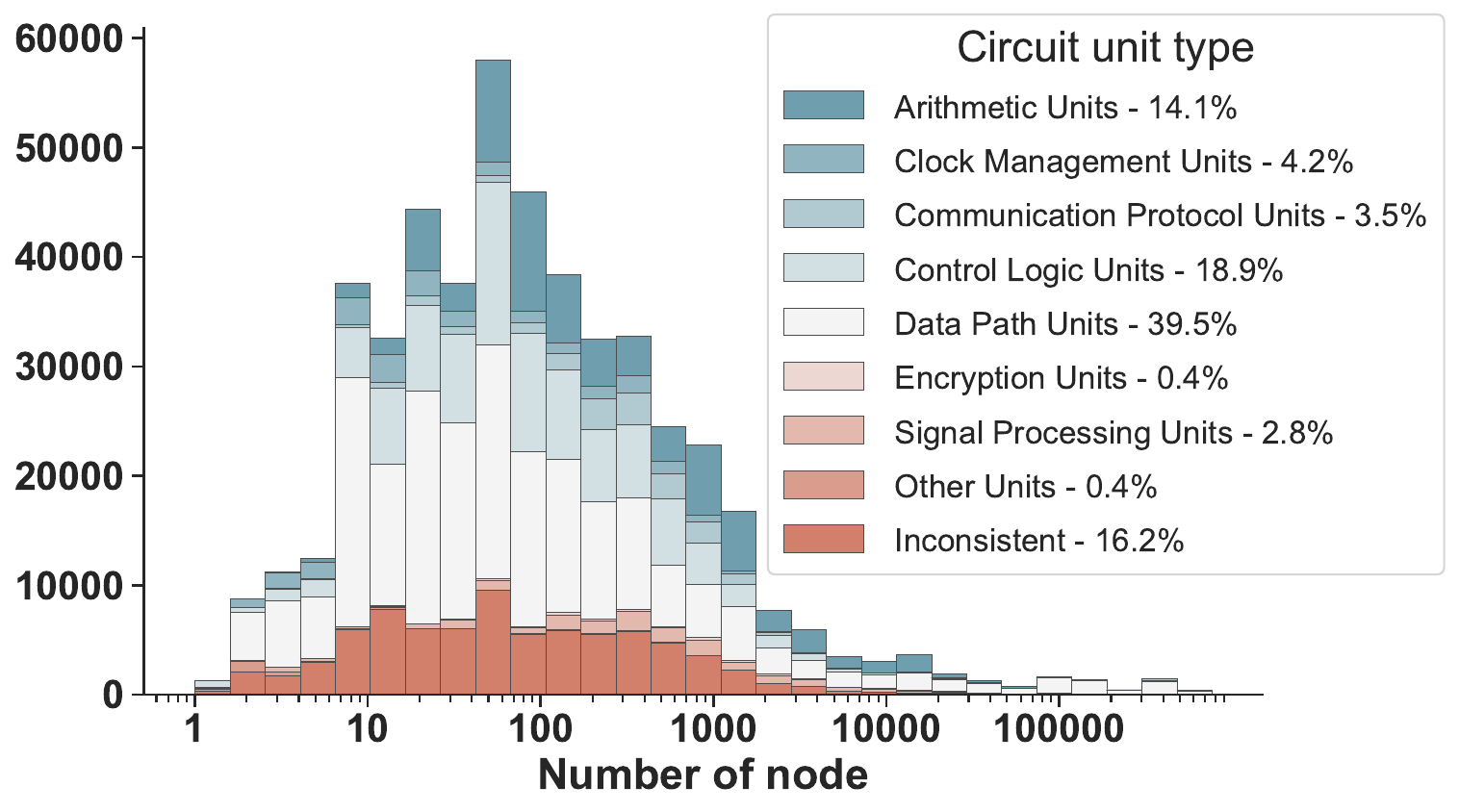}
    \caption{Histogram of node count for netlist graphs in $\mathsf{BRIDGES}$.
    The stacked bars represent different circuit type labels, while Inconsistent represents different circuit-type label predictions between LLaMA-3-70B and GPT-4o.
    }
    \label{fig:histogram_dataset}
    \vspace{-4mm}
\end{figure}

A general overview of the dataset is shown in \Cref{fig:data_example} (Right).
For text modality, $\mathsf{BRIDGES}$ contains 37,676 RTL designs, 744,768 netlists from logic synthesis, function descriptions, and their circuit types.
For graph modality, $\mathsf{BRIDGES}$ includes 32,569 dataflow graphs for the RTL, and 503,362 netlist graphs.
The histogram in \Cref{fig:histogram_dataset} illustrates the distribution of the node count for netlist graphs in $\mathsf{BRIDGES}$. 
Though most netlist graphs have fewer than 1,000 nodes, there are approximately 10,000 graphs with over 10,000 nodes, with a maximum node count of nearly 800,000.
\Cref{tab:dataset_comparison} compares $\mathsf{BRIDGES}$ with other datasets. 
$\mathsf{BRIDGES}$ is the first large-scale dataset in EDA to provide both text and graph modalities, encompassing 500K netlist and dataflow graphs, 
with a total of more than 1.5 billion tokens. 
Compared to PubChem \cite{liu2023molca,cao2023instructmol}, a widely used molecular graph dataset in chemistry domain, $\mathsf{BRIDGES}$ has significantly more tokens, 
reflecting the complexity of netlist graphs.
However, when compared to MINT-1T \cite{awadalla2024mint}, the state-of-the-art large-scale dataset in the vision domain, $\mathsf{BRIDGES}$ contains substantially fewer tokens. As demonstrated in \Cref{sec:scale}, 
the current dataset size limits model performance, indicating the potential for better performance with larger datasets. 

\begin{table}[]
    \centering
    \resizebox{.5\textwidth}{!}{\begin{tabular}{l|l|l|l}
    \toprule
        \textbf{Datasets} & \textbf{Domain} & \textbf{Modality}   &  \textbf{No. of tokens} \\
    \midrule
        $\mathsf{BRIDGES}$ & EDA & graph, text  & 1.5B$^1$ \\
        RTLCoder \cite{liu2024rtlcoder} & EDA & text & 1.3M \\
        MG-Verilog \cite{zhang2024mg} & EDA & text & 0.5M \\
        PubChem \cite{liu2023molca,cao2023instructmol} & Chemistry & graph, text & 10M$^1$ \\
        MINT-1T \cite{awadalla2024mint} & Vision & image, text & 1T \\
    \bottomrule
    \end{tabular}}
    \caption{Comparison of $\mathsf{BRIDGES}$ with other datasets.}
    \label{tab:dataset_comparison}
    \begin{tablenotes}
        \footnotesize
        \item 1: The token count for graph modality is calculated using the total number of nodes, following the approach described in \cite{chai2023graphllm}.
    \end{tablenotes}
    \vspace{-4mm}
\end{table}

\section{Model Architecture}
\begin{figure*}
    \centering
    \includegraphics[width=1.05\textwidth]{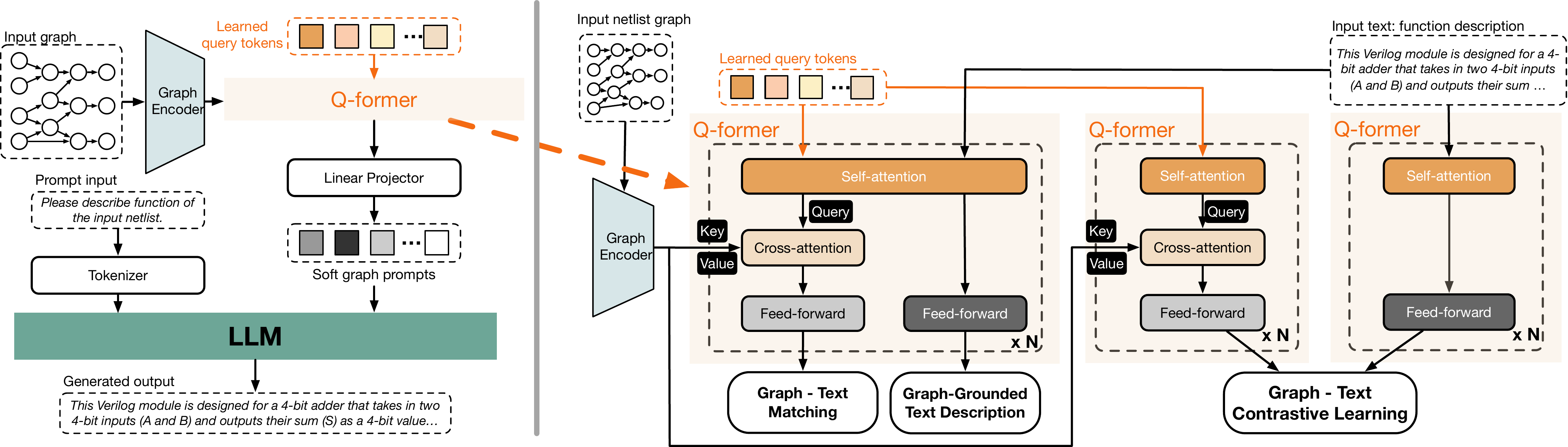}
    \caption{\textbf{Left}: The graph-supported LLM architecture in $\mathsf{BRIDGES}$.
    \textbf{Right}: Stage 1 pre-training of $\mathsf{BRIDGES}$. 
    The graph encoder and the cross-modal projector (Q-Former) are optimized together through three cross-modal tasks. 
    Modules with the same color share the same parameters.}
    \label{fig:model_arch}
    \vspace{-4mm}
\end{figure*}

In this section, we introduce the model architecture, that bridges the modality gap and brings the graph modality into the LLMs.
As shown in \Cref{fig:model_arch}, the architecture is composed of three key components:
1) a graph encoder, which encodes the graph structure in EDA tasks, such as netlist graph or RTL dataflow graph.
2) a cross-modal projector based on Querying Transformer (Q-Former) module \cite{li2023blip}. It bridges the gap between the text and graph modality.
3) a LLM that takes the graph information from the cross-modal projector and the text input, 
and generates the output sequence.
We describe the details of each component in the following sections.

\subsection*{Graph encoder}
A graph encoder outputs a fixed-size representation for the graph, and leverages the rich structural and function information intrinsic to the graph.
The architecture of a graph encoder can be modulated depending on the specific EDA tasks and graph types.
However, graph encoders used in other domains \cite{liu2023molca,cao2023instructmol, chai2023graphllm}
are not directly applicable to EDA tasks.
For example, the query vector in MolCA \cite{liu2023molca} attends to each node embedding in the graph, 
GraphLLM \cite{chai2023graphllm} uses a graph transformer to encode the graph structure.
The large size of typical graphs in EDA tasks makes these techniques infeasible.

In this work, we adopt NetlistGNN \cite{li2023characterize} as the base structure of our graph encoder to encode the netlist graph.
NetlistGNN is a message-passing-based graph neural network that represents the netlist graph as a directed heterogeneous graph.
After NetlistGNN obtains the node embeddings, we apply a series of predefined pooling operations \texttt{Pool} to obtain the graph embedding.
In the experiments, $\texttt{Pool}$ includes mean, max, sum, and min pooling.
Formally, given a graph $\gG = (\gV, \gE)$, where $\gV$ is the set of nodes and $\gE$ is the set of edges,
the graph encoder can be formulated as:
\begin{align}
    \mathbf{h}_{\gV} & = \text{NetlistGNN}(\gG) \in \R^{|\gV| \times d_1} \nonumber\\
    \mathbf{h}_{\gG} & = \oplus_{i=1}^{|\texttt{Pool}|} \texttt{Pool}_i(\mathbf{h}_{\gV}) \in \R^{d_1 \times |\texttt{Pool}|}
\end{align}
where $\mathbf{h}_{\gV}$, $\mathbf{h}_{\gG}$ are the node embeddings and the graph embedding, respectively, 
$d_1$ is the dimension of the node embeddings,  $\oplus$ is the stack operation, and $\texttt{Pool}_i$ is the $i$-th pooling operation.
The graph embedding $\mathbf{h}_{\gG}$ is then fed into the cross-modal projector as keys and values in the cross-attention module.

We emphasize that while the graph encoder is not the primary focus of this work, its significance should not be overlooked. 
In fact, our experiments in \Cref{sec:scale} demonstrate that a larger graph encoder exhibits superior representation capabilities, resulting in improved performance on the design retrieval task. Investigating more powerful graph encoders remains an avenue for future research.

\subsection*{Cross-modal projector}

The cross-modal projector is a bridge that enables a LLM to capture graph information.
We use the Querying Transformer (Q-Former) \cite{li2023blip} as the base structure.
As shown in \Cref{fig:model_arch} (right),
the Q-Former receives a set of learnable query tokens as input, with the query numbers $q$ treated as model parameters. 
These queries engage with one another via self-attention modules and connect with 
graph embeddings through cross-attention modules, where graph embeddings function as keys and values.
These cross-attention modules are added to every alternate transformer block and in total 12 transformer blocks are used in the experiments. 
Let $\vq \in \R^{q \times d_2}$ be the query tokens and $d_2$ is the dimension of the query tokens.
The cross-modal projector can be formulated as:
\begin{align}
    \mathbf{h}_{\vq} & = \text{Q-Former}(\vq, \mathbf{h}_{\gG}) \in \R^{q \times d_3}
\end{align}
where $\mathbf{h}_{\vq}$ is the output of the cross-modal projector with dimension $d_3$.

\subsection*{Base LLM}
We adopt Llama3 \cite{dubey2024llama} as our foundation LLM, while our method is not limited to Llama3 but applicable to any LLM.
Specifically, we project the cross-modal output $\mathbf{h}_{\vq}$ to 
match the dimensionality of text token embeddings, 
effectively serving as soft graph prompt tokens for the input sequence.

\section{Training Strategy}
The training in $\mathsf{BRIDGES}$ includes two stages: 
stage 1 is the pre-training representation learning stage to extract text-relevant graph representation,
while stage 2 is the alignment learning stage to align the graph representation with the selected LLM for a specific task.

 
\subsection{Stage 1 - Pre-training as representation learning}

In stage 1, we connect graph encoder to Q-Former and perform pre-training using graph-text pairs. 
Here, we use the netlist graph and the function description as pairs.
It should be noted that the method is general and can be applied to other graph-text pairs, e.g., RTL dataflow graph and RTL code.
 The goal is for the queries in Q-Former to be able to extract graph representations that are most informative for the text. 
 Inspired by BLIP2 \cite{li2023blip}, which demonstrates success in aligning vision and language modalities,
  we jointly optimize three pre-training objectives that share the same input format and model parameters. 
As shown in \Cref{fig:model_arch} (right), the three objectives are:

\subsubsection*{Graph-text contrastive learning}
Graph-Text Contrastive Learning (GTC) maximizes mutual information between graph and text representations by contrasting the similarity of positive graph-text pairs with negative ones (see \Cref{fig:model_arch}, right). 
Specifically, we sample a positive graph-text pair and a negative graph-text pair from the same batch.
For each pair, query tokens and text tokens are fed into the Q-Former separately, without attending to each other.
For the query tokens, the graph embedding from the graph encoder is used as the key and value in the cross-attention module, 
while the text tokens only involve self-attention.
The output query representation from Q-former $\vh_q$ is aligned with the output text representation (also from Q-former) by a contrastive loss.
Especially, when computing the similarity between $\vh_q$ and the output text representation, the largest similarity among all query tokens is used as the final similarity score.

\subsubsection*{Graph-text matching}
Graph-text matching (GTM) is a binary classification task that learns fine-grained graph-text alignment. 
The model predicts whether a graph-text pair is matched or unmatched.
Specifically, the query tokens and text tokens are concatenated and fed into the Q-Former together, where they attend to each other, allowing full query-text interaction, i.e., bidirectional communication between query embeddings and text embeddings during the attention process.
The output query embeddings $\vh_q$ are passed into a two-class linear classifier to generate logits, 
with the final matching score averaged across all queries. 
Within a batch, the negative pairs with the highest similarity scores (in contrastive learning) are selected as hard negatives for the matching task.

\subsubsection*{Graph-grounded text generation}
Graph-grounded text generation (GTG) trains the Q-Former to generate text conditioned on graphs. 
Similar with GTM, the query tokens and text tokens are concatenated and fed into the Q-Former together.
However, in GTG, query tokens attend to each other but not text tokens, and each text token can attend to all queries and previous text tokens. 
Text tokens can only access graph information via the queries, compelling the queries to extract graph knowledge through the cross-attention modules.

\subsection{Stage 2 - Alignment learning}
In this stage, the goal is to align outputs of the cross-modal projector, $\vh_q$,  with the adopted LLM, and to utilize the generative language capabilities of the LLM.
As shown in \Cref{fig:model_arch}, a linear projector layer projects the output query embeddings 
$\vh_q$ to match the dimensionality of the LLM text embeddings. 
These projected embeddings are prepended to the input text embeddings, functioning as soft graph prompts that condition the LLM on the graph information extracted by Q-Former. 
Compared with traditional \textit{Graph2Text}, the graph information is encoded in the query tokens, whose number is much smaller than the number of tokens in \textit{Graph2Text}, and helps prevent catastrophic forgetting.
The training for stage 2 is task-specific in which we collect the task-specific data instance and follow the traditional LLM training pipeline, 
where the model generates the output sequence token by token.

\section{Experiments}
\label{sec:result}
Our experiments evaluate  $\mathsf{BRIDGES}$ across four tasks.
Additionally, we conduct a comprehensive scalability study to assess the impact of model and data scale, and perform an extensive ablation study to determine the contributions of various components in $\mathsf{BRIDGES}$, including modality, LLM fine-tuning, pre-training, and the cross-modal projector.

\label{sec:scale}
\begin{figure*}
    \centering
    \includegraphics[width=1.02\textwidth]{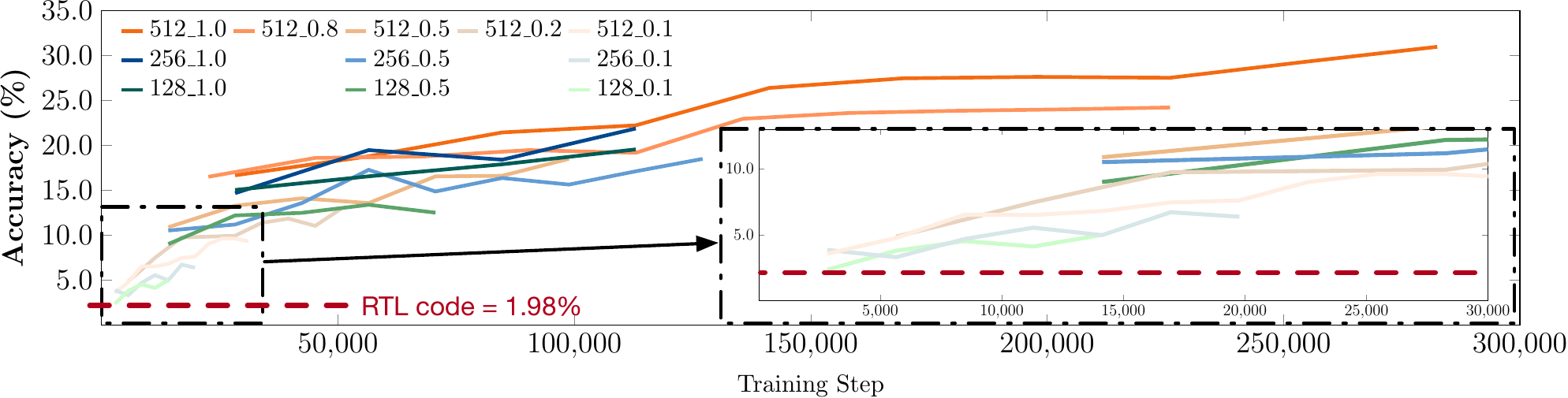}
    \caption{Accuracy of \textit{Function2Design} on the full test set (25,569 graphs) with varying model and data scales. 
    Training is stopped early if the accuracy on the validation set decreases for two consecutive epochs.
    The legend $d_1\_x$ denotes the dimension of node embeddings in the graph encoder, and $x \times 100\%$ represents the proportion of training data used. The red dotted line shows the accuracy (1.98\%) when using RTL code to represent the design instead of a netlist graph.
    }
    \label{fig:scalability}
    \vspace{-4mm}
\end{figure*}
\subsection{Experiment setting}
We partition the collected dataset into training, validation, and test sets in an 18:1:1 ratio based on RTL code, ensuring that no design overlaps across sets. 
The training set contains 452,050 netlist graphs, with 25,743 and 25,569 graphs in the validation and test sets, respectively.
We anonymize the module names in RTL code and netlist code to prevent LLMs from easily guessing.
The training process is conducted in two stages: the first stage runs for 10 epochs, and the second for 3 epochs. 
NetlistGNN \cite{li2023characterize}, the graph encoder used in $\mathsf{BRIDGES}$, consists of five layers, each with a dimension of 512.
Q-Former uses $\text{BERT}_{\text{base}}$ \cite{devlin2018bert} as the base architecture and loads its pre-trained weights, 
while the cross-attention layers start with random initialization. 
The number of query tokens is 8 ($q$ = 8). The optimizer configuration follows BLIP2 \cite{li2023blip} as it has shown effectiveness in vision-language tasks.
In the experiments, we use Llama herd \cite{dubey2024llama} as the base LLMs. ``3B(1B)" and ``Llama3-3B(1B)" refer to Llama-3.2-3B(1B)-instruct, while ``8B" and ``Llama3-8B" denote Llama-3.1-8B-instruct. 
The max sequence length in set to 2048, we observe that text-only LLMs show even worse performance with longer max sequence length.
This limitation likely stems from catastrophic forgetting when processing lengthy text representations generated by \textit{Graph2Text}.
All experiments are performed on a single 80GB NVIDIA H100 GPU. Training $\mathsf{BRIDGES}$ for each task on this hardware is completed within two days.


\begin{table}[t!]
    \small
    \centering
    \begin{tabular}{cccccc} \toprule
Retrieval                                     & {Design }                 & \multicolumn{2}{c}{\textit{Design2Function}} & \multicolumn{2}{c}{\textit{Function2Design}} \\
type                                          &    representation                                           & Acc                   & R@20                 & Acc                   & R@20                 \\  \midrule
    \multicolumn{1}{c|}{\multirow{3}{*}{In-batch$^1$}} & \multicolumn{1}{c|}{Netlist (text)}           & 5.39                  & 46.45                & 4.49                  & 46.85                \\
    \multicolumn{1}{c|}{}                         & \multicolumn{1}{c|}{RTL (text)}               & 16.09                 & 65.80                & 12.15                 & 67.79                \\
    \multicolumn{1}{c|}{}                         & \multicolumn{1}{c|}{\textbf{Netlist (graph)}} & \textbf{63.77}        & \textbf{93.04}       & \textbf{63.84}        & \textbf{93.07}       \\ \hline
    \multicolumn{1}{c|}{\multirow{3}{*}{Fullset$^2$}}        & \multicolumn{1}{c|}{Netlist (text)}           & 1.26                  & 1.27                 & 0.42                  & 0.63                 \\
    \multicolumn{1}{c|}{}                         & \multicolumn{1}{c|}{RTL (text)}               & 3.59                  & 3.59                 & 1.98                  & 1.98                 \\
    \multicolumn{1}{c|}{}                         & \multicolumn{1}{c|}{\textbf{Netlist (graph)}} & \textbf{30.56}        & \textbf{30.59}       & \textbf{30.98}        & \textbf{42.83}       \\ \bottomrule
    \end{tabular}
    \caption{Results of design retrieval.  In-batch refers to retrieval in a batch of 64 random samples. Fullset refers to retrieval in the full test set (25,569 graphs). 
     }
    \label{tab:retrieval}
    \vspace{-4mm}
\end{table}

\subsection{Design retrieval}
Efficiently locating existing designs within extensive databases is helpful to reduce redundant re-implementation and enhance design reuse.
To substantiate our claim that text-only approaches have inherent limitations in EDA, we construct a database using test set,
where designs and their function descriptions are stored as embeddings. 
The function descriptions are encoded using $\text{BERT}_{\text{base}}$, 
while design embeddings are derived from Q-former's query embeddings.
For text-based design representation (netlist files and RTL code), a separate $\text{BERT}_{\text{base}}$ replaces the graph encoder to connect Q-former.
The retrieval process involves computing similarity scores between an input embedding and all embeddings in the database, with the highest-scoring match being selected. 
We assess retrieval performance across three distinct design representations: netlist graph, netlist file, and RTL code, 
with each representation trained independently. 
The assessment covers both retrieval directions: retrieving designs using function embedding (\textit{Function2Design}) and retrieving function descriptions using design embedding (\textit{Design2Function}).
\Cref{tab:retrieval} presents the results, demonstrating that netlist graphs significantly outperform the text-based inputs:
netlist file representation yields only marginally better results than random guessing,
while using RTL code improves performance, it remains far inferior to netlist graph (lower by 47\% in in-batch accuracy and 27\% in fullset accuracy), despite RTL being a higher-level abstraction that theoretically contains more information. 

\subsection{Type classification}
\label{sec:type}
\begin{figure}
    \centering
    \includegraphics[width=.45\textwidth]{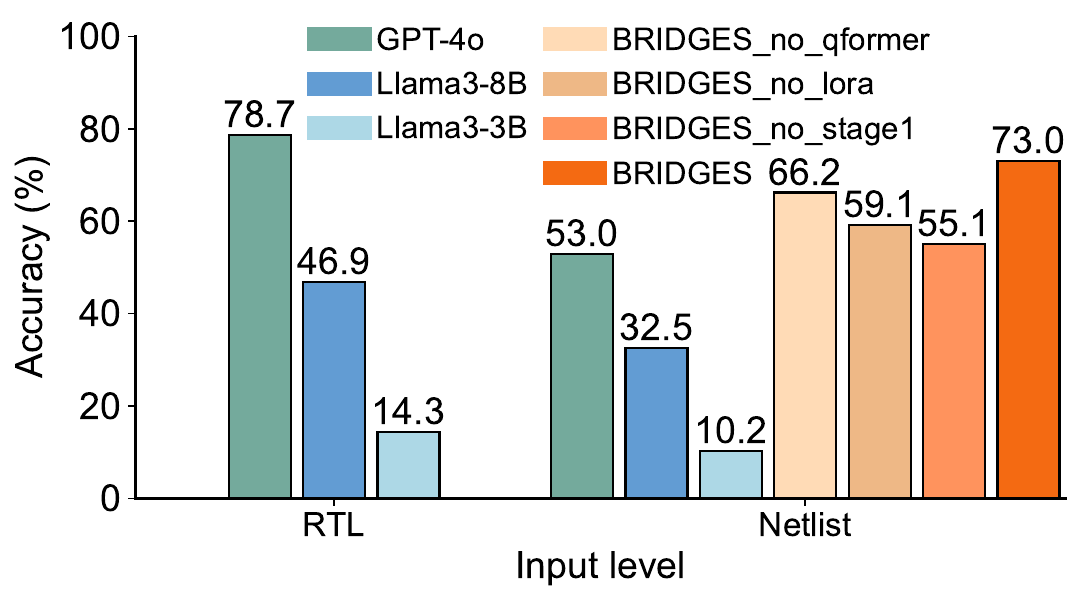}
    \caption{The type prediction accuracy of $\mathsf{BRIDGES}$-3B (orange bars), Llama3 herd (blue bars), and GPT-4o (green).}
    \label{fig:type}
\end{figure}

We assess the performance of $\mathsf{BRIDGES}$ in classifying circuit types, a critical prerequisite for enabling AI agents to effectively reason about circuit designs. 
We first use function description to query the type from GPT-4o and Llama3-70B, with their response as the ground truth.
Data instances with inconsistent labels between the two models are excluded to ensure label reliability.
For $\mathsf{BRIDGES}$, we input netlist graphs along with corresponding questions into the model to predict circuit types. 
As a baseline, we evaluate Llama3 and GPT-4o using either netlist text or RTL code as input. 
Results are summarized in \Cref{fig:type}.
In netlist-level comparisons, $\mathsf{BRIDGES}$, equipped with Llama3-3B fine-tuned via LoRA \cite{hu2021lora}, 
achieves a 40.5\% accuracy improvement over Llama3-8B, 62.8\% over text-only Llama3-3B, 
and outperforms GPT-4o by 20\%. When compared with LLMs using RTL code as input, 
$\mathsf{BRIDGES}$ still surpasses Llama3-8B by 26.1\% but slightly underperforms GPT-4o by 5.7\%. 
This discrepancy arises because RTL code inherently provides higher-level abstractions, 
while $\mathsf{BRIDGES}$ accepts solely netlist graphs as input.


\subsection{Function description generation}
We evaluate $\mathsf{BRIDGES}$ on its ability to describe the function of a given design, 
using perplexity as the performance metric. Perplexity quantifies how effectively a model predicts a text sample (in this case, a function description), with lower values indicating better performance.
As shown in \Cref{tab:func_des_result}, $\mathsf{BRIDGES}$ with a 3B LLM achieves a perplexity of 2.08 when provided with both netlist graphs and RTL code 
as input. In contrast, Llama3.1-8B, the best-performing model from the Llama herd in this experiment, records a perplexity of 4.65 for RTL text input. These results highlight the superior capability of $\mathsf{BRIDGES}$ in understanding designs and generating accurate function descriptions.

\begin{table}[]
    \small
\centering
\resizebox{.5\textwidth}{!}{\begin{tabular}{ccc|c|cc} \toprule
    \multicolumn{3}{c|}{Llama3 (text only)} & \multirow{2}{*}{Graph-only} & \multicolumn{2}{c}{$\mathsf{BRIDGES}$-3B} \\
    1B        & 3B        & 8B        &                             & \textcolor{myorange}{Fine tuning} LLM       & \textcolor{lightblue}{Freeze} LLM      \\ \midrule
    6.07      & 5.16      & 4.65      & \multicolumn{1}{c|}{2.53}   & \textbf{2.08  }       & 2.23           \\ \bottomrule 
    \end{tabular}}
\caption{Perplexity (PPL) for function description generation. 
Llama3 refer to representing design as RTL code (text modality only). 
Graph-only means only using graph encoder and Q-former for generation (graph modality only).
BRIDGES takes both modalities as input.}
\label{tab:func_des_result}
\end{table}

\subsection{Area and power estimation}
Here, GPT-4o and $\mathsf{BRIDGES}$ are asked to predict the area and power of provided designs.
GPT-4o is provided with few-shot examples, and
$\mathsf{BRIDGES}$ uses a training-free retrieval-augmented method \cite{fang2025circuitfusion}, 
where the most similar design is retrieved by $\mathsf{BRIDGES}$ as a reference for estimation.
Additionally, we use the same retrieval method, but with text modality only (``Retrieval-text-only") as another baseline.
As shown in \Cref{tab:ppa}, $\mathsf{BRIDGES}$ achieves a smaller than $1\%$ MAPE for over $75\%$ of designs on both area and power estimation,
significantly outperforming GPT-4o and text-only retrieval approach.

\begin{table}[]
    \small
\centering
\begin{tabular}{l|ll|ll|ll} \toprule
    \multicolumn{1}{c|}{\multirow{2}{*}{MAPE}} & \multicolumn{2}{c|}{GPT-4o} & \multicolumn{2}{c|}{Retrival-text-only} & \multicolumn{2}{c}{$\mathsf{BRIDGES}$} \\
    \multicolumn{1}{c|}{}                      & Area         & Power        & Area                & Power                & Area         & Power        \\ \midrule
    \textless{}1\%                             & 0.013        & 0            & 0.023                    &  0.021                    & \textbf{0.761}           & \textbf{0.756}           \\
    \textless{}10\%                            & 0.045        & 0.024        & 0.106                 &   0.097                  & \textbf{0.796 }             & \textbf{0.786}             \\
    \textless{}50\%                            & 0.261        & 0.205        & 0.188                    &  0.182                   &  \textbf{0.859}            & \textbf{0.855}            \\ \bottomrule
    \end{tabular}

\caption{Area and power estimation results. $<x\%$ means the percentage of Mean Absolute Percentage Error (MAPE) less than $x\%$.
}
\label{tab:ppa}
\end{table}

\subsection{Scalability study}
Scalability is a key factor contributing to the success of LLMs. Here, we examine the scalability of $\mathsf{BRIDGES}$ with respect to both model size and data size. The results are summarized in \Cref{fig:scalability}.
We observe that graph encoders with smaller node embedding sizes, such as $d_1 = 256$ (blue curves) and $d_1 = 128$ (green curves), reach lower accuracy ceilings compared to $d_1 = 512$ and exhibit signs of over-fitting as the training dataset grows. In contrast, the $d_1 = 512$ configuration (orange curves) demonstrates consistent performance improvement with increasing data size, without signs of saturation, suggesting that this model remains data-limited.

\subsection{Runtime and memory overhead}
\label{sec:runtime}
\begin{figure}
    \centering
    \includegraphics[width=.35\textwidth]{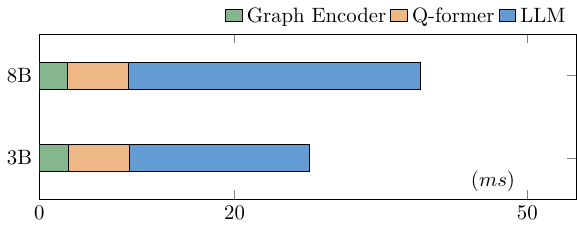}
    \caption{The average forward time per step of $\mathsf{BRIDGES}$ with 3B and 8B LLMs for a batch size = 8.}
    \label{fig:runtime}
    \vspace{-4mm}
\end{figure}

The runtime distribution is detailed in \Cref{fig:runtime}. The additional runtime overhead introduced by $\mathsf{BRIDGES}$, including the graph encoder and Q-former, accounts for only approximately 30\% and 20\% for 3B and 8B LLMs, respectively. This overhead is negligible compared to the significant performance gains demonstrated in previous sections.
The additional memory cost is even less significant. The 512-dimensional graph encoder comprises 5.3M trainable parameters, and the Q-former includes 96.9M parameters. In contrast, the LLMs are hundreds of times larger, rendering these contributions minimal in comparison.

\subsection{Ablation study}
We conduct ablation study to assess the impact of factors we are concerned with in $\mathsf{BRIDGES}$.

\noindent \textbf{Fine-tuning LLM} As shown in \Cref{fig:type}, freezing the LLM during stage 2 (``$\mathsf{BRIDGES}$\_no\_lora")—where only the graph encoder and Q-former are fine-tuned—reduces accuracy 
from 73.0\% to 59.1\%. 
However, this reduced performance still substantially outperforms baseline models, 
achieving nearly double the accuracy of Llama3-8B and four times that of Llama3-3B.
A similar trend is observed in function description generation (see \Cref{tab:func_des_result}):
freezing the LLM increases perplexity from 2.08 to 2.23, 
while still maintaining significantly better results than the Llama models. 
These findings underscore the efficacy of graph prompt tokens in enhancing the LLM's understanding of designs,
without requiring fine-tuning of the model, offering a cost-effective approach to improving performance.

\noindent \textbf{Pre-training}
The omission ("$\mathsf{BRIDGES}$\_no\_stage1") of pre-training (stage 1) results in a further accuracy decline to 55.1\% in type prediction task (see \Cref{fig:type}). 
This degradation demonstrates the effectiveness of pre-training for down-stream tasks.
Nevertheless, the performance without pre-training remains significantly superior to text-only LLMs, i.e., 5x better than same-scale Llama3-3B.

\noindent \textbf{Graph modality only} To show the advantage of combining different modalities, we evaluate the performance of $\mathsf{BRIDGES}$ against text-only LLMs and graph-only model (graph encoder + Q-former) in \Cref{tab:func_des_result}.
The graph-only model achieves a perplexity of 2.53, outperforming text-only LLMs but remains worse than $\mathsf{BRIDGES}$ (2.08).
The superior performance of BRIDGES highlights the advantage of combining text and graph modalities, allowing the LLM to utilize complementary strengths from both representations. 

\noindent \textbf{Cross-modal projector} We also access the impact of Q-former on the performance of $\mathsf{BRIDGES}$ by replacing it with a simple cross-modal projector (linear layer) to project the graph embedding to the text space.
As shown in \Cref{fig:type}, the accuracy drops from 73.0\% to 66.2\%, indicating that the Q-former is crucial for effectively bridging the gap between the two modalities.

\section{Conclusion}
\label{sec:conclu}
In this work, we propose $\mathsf{BRIDGES}$, a framework designed to incorporate graph modality into LLMs for EDA tasks.
$\mathsf{BRIDGES}$ is composed of an automated data generation workflow, 
a solution to bridge graph and text modalities in LLMs.
Experimental results demonstrate its effectiveness and superior performance over text-only LLMs and graph-only models.
Looking forward, the study of graph encoder deserves more exploration.



\clearpage
{
\bibliographystyle{IEEEtran}
\bibliography{main.bbl}

\begin{thebibliography}{10}
\providecommand{\url}[1]{#1}
\csname url@samestyle\endcsname
\providecommand{\newblock}{\relax}
\providecommand{\bibinfo}[2]{#2}
\providecommand{\BIBentrySTDinterwordspacing}{\spaceskip=0pt\relax}
\providecommand{\BIBentryALTinterwordstretchfactor}{4}
\providecommand{\BIBentryALTinterwordspacing}{\spaceskip=\fontdimen2\font plus
\BIBentryALTinterwordstretchfactor\fontdimen3\font minus
  \fontdimen4\font\relax}
\providecommand{\BIBforeignlanguage}[2]{{%
\expandafter\ifx\csname l@#1\endcsname\relax
\typeout{** WARNING: IEEEtran.bst: No hyphenation pattern has been}%
\typeout{** loaded for the language `#1'. Using the pattern for}%
\typeout{** the default language instead.}%
\else
\language=\csname l@#1\endcsname
\fi
#2}}
\providecommand{\BIBdecl}{\relax}
\BIBdecl

\bibitem{brown2020language}
T.~B. Brown, ``Language models are few-shot learners,'' \emph{arXiv preprint
  arXiv:2005.14165}, 2020.

\bibitem{bubeck2023sparks}
S.~Bubeck, V.~Chandrasekaran, R.~Eldan, J.~Gehrke, E.~Horvitz, E.~Kamar,
  P.~Lee, Y.~T. Lee, Y.~Li, S.~Lundberg \emph{et~al.}, ``Sparks of artificial
  general intelligence: Early experiments with gpt-4,'' \emph{arXiv preprint
  arXiv:2303.12712}, 2023.

\bibitem{wei2022chain}
J.~Wei, X.~Wang, D.~Schuurmans, M.~Bosma, F.~Xia, E.~Chi, Q.~V. Le, D.~Zhou
  \emph{et~al.}, ``Chain-of-thought prompting elicits reasoning in large
  language models,'' \emph{Advances in neural information processing systems},
  vol.~35, pp. 24\,824--24\,837, 2022.

\bibitem{liu2024rtlcoder}
S.~Liu, W.~Fang, Y.~Lu, J.~Wang, Q.~Zhang, H.~Zhang, and Z.~Xie, ``Rtlcoder:
  Fully open-source and efficient llm-assisted rtl code generation technique,''
  \emph{IEEE Transactions on Computer-Aided Design of Integrated Circuits and
  Systems}, 2024.

\bibitem{pei2024betterv}
Z.~Pei, H.-L. Zhen, M.~Yuan, Y.~Huang, and B.~Yu, ``Betterv: Controlled verilog
  generation with discriminative guidance,'' \emph{arXiv preprint
  arXiv:2402.03375}, 2024.

\bibitem{chang2024data}
K.~Chang, K.~Wang, N.~Yang, Y.~Wang, D.~Jin, W.~Zhu, Z.~Chen, C.~Li, H.~Yan,
  Y.~Zhou \emph{et~al.}, ``Data is all you need: Finetuning llms for chip
  design via an automated design-data augmentation framework,'' \emph{arXiv
  preprint arXiv:2403.11202}, 2024.

\bibitem{chang2024natural}
K.~Chang, Z.~Chen, Y.~Zhou, W.~Zhu, H.~Xu, C.~Li, M.~Wang, S.~Liang, H.~Li,
  Y.~Han \emph{et~al.}, ``Natural language is not enough: Benchmarking
  multi-modal generative ai for verilog generation,'' \emph{arXiv preprint
  arXiv:2407.08473}, 2024.

\bibitem{xu2024meic}
K.~Xu, J.~Sun, Y.~Hu, X.~Fang, W.~Shan, X.~Wang, and Z.~Jiang, ``Meic:
  Re-thinking rtl debug automation using llms,'' \emph{arXiv preprint
  arXiv:2405.06840}, 2024.

\bibitem{yao2024rtlrewriter}
X.~Yao, Y.~Wang, X.~Li, Y.~Lian, R.~Chen, L.~Chen, M.~Yuan, H.~Xu, and B.~Yu,
  ``Rtlrewriter: Methodologies for large models aided rtl code optimization,''
  \emph{arXiv preprint arXiv:2409.11414}, 2024.

\bibitem{wu2024chateda}
H.~Wu, Z.~He, X.~Zhang, X.~Yao, S.~Zheng, H.~Zheng, and B.~Yu, ``Chateda: A
  large language model powered autonomous agent for eda,'' \emph{IEEE
  Transactions on Computer-Aided Design of Integrated Circuits and Systems},
  2024.

\bibitem{pu2024customized}
Y.~Pu, Z.~He, T.~Qiu, H.~Wu, and B.~Yu, ``Customized retrieval augmented
  generation and benchmarking for eda tool documentation qa,'' \emph{arXiv
  preprint arXiv:2407.15353}, 2024.

\bibitem{guo2020graphcodebert}
D.~Guo, S.~Ren, S.~Lu, Z.~Feng, D.~Tang, S.~Liu, L.~Zhou, N.~Duan,
  A.~Svyatkovskiy, S.~Fu \emph{et~al.}, ``Graphcodebert: Pre-training code
  representations with data flow,'' \emph{arXiv preprint arXiv:2009.08366},
  2020.

\bibitem{li2022rethinking}
W.~Li, R.~Li, Y.~Ma, S.~O. Chan, D.~Pan, and B.~Yu, ``Rethinking graph neural
  networks for the graph coloring problem,'' \emph{arXiv preprint
  arXiv:2208.06975}, 2022.

\bibitem{wang2024can}
H.~Wang, S.~Feng, T.~He, Z.~Tan, X.~Han, and Y.~Tsvetkov, ``Can language models
  solve graph problems in natural language?'' \emph{Advances in Neural
  Information Processing Systems}, vol.~36, 2024.

\bibitem{chai2023graphllm}
Z.~Chai, T.~Zhang, L.~Wu, K.~Han, X.~Hu, X.~Huang, and Y.~Yang, ``Graphllm:
  Boosting graph reasoning ability of large language model,'' \emph{arXiv
  preprint arXiv:2310.05845}, 2023.

\bibitem{liu2024lost}
N.~F. Liu, K.~Lin, J.~Hewitt, A.~Paranjape, M.~Bevilacqua, F.~Petroni, and
  P.~Liang, ``Lost in the middle: How language models use long contexts,''
  \emph{Transactions of the Association for Computational Linguistics},
  vol.~12, pp. 157--173, 2024.

\bibitem{zhang2024mg}
Y.~Zhang, Z.~Yu, Y.~Fu, C.~Wan, and Y.~C. Lin, ``Mg-verilog: Multi-grained
  dataset towards enhanced llm-assisted verilog generation,'' in \emph{2024
  IEEE LLM Aided Design Workshop (LAD)}.\hskip 1em plus 0.5em minus 0.4em\relax
  IEEE, 2024, pp. 1--5.

\bibitem{thakur2023autochip}
S.~Thakur, J.~Blocklove, H.~Pearce, B.~Tan, S.~Garg, and R.~Karri, ``Autochip:
  Automating hdl generation using llm feedback,'' \emph{arXiv preprint
  arXiv:2311.04887}, 2023.

\bibitem{chang2023chipgpt}
K.~Chang, Y.~Wang, H.~Ren, M.~Wang, S.~Liang, Y.~Han, H.~Li, and X.~Li,
  ``Chipgpt: How far are we from natural language hardware design,''
  \emph{arXiv preprint arXiv:2305.14019}, 2023.

\bibitem{blocklove2023chip}
J.~Blocklove, S.~Garg, R.~Karri, and H.~Pearce, ``Chip-chat: Challenges and
  opportunities in conversational hardware design,'' in \emph{2023 ACM/IEEE 5th
  Workshop on Machine Learning for CAD (MLCAD)}.\hskip 1em plus 0.5em minus
  0.4em\relax IEEE, 2023, pp. 1--6.

\bibitem{liu2023verilogeval}
M.~Liu, N.~Pinckney, B.~Khailany, and H.~Ren, ``Verilogeval: Evaluating large
  language models for verilog code generation,'' in \emph{2023 IEEE/ACM
  International Conference on Computer Aided Design (ICCAD)}.\hskip 1em plus
  0.5em minus 0.4em\relax IEEE, 2023, pp. 1--8.

\bibitem{xiong2024hlspilot}
C.~Xiong, C.~Liu, H.~Li, and X.~Li, ``Hlspilot: Llm-based high-level
  synthesis,'' \emph{arXiv preprint arXiv:2408.06810}, 2024.

\bibitem{yin2024ado}
Y.~Yin, Y.~Wang, B.~Xu, and P.~Li, ``Ado-llm: Analog design bayesian
  optimization with in-context learning of large language models,'' \emph{arXiv
  preprint arXiv:2406.18770}, 2024.

\bibitem{fang2025circuitfusion}
\BIBentryALTinterwordspacing
W.~Fang, S.~Liu, J.~Wang, and Z.~Xie, ``Circuitfusion: Multimodal circuit
  representation learning for agile chip design,'' in \emph{The Thirteenth
  International Conference on Learning Representations}, 2025. [Online].
  Available: \url{https://openreview.net/forum?id=rbnf7oe6JQ}
\BIBentrySTDinterwordspacing

\bibitem{cao2023instructmol}
H.~Cao, Z.~Liu, X.~Lu, Y.~Yao, and Y.~Li, ``Instructmol: Multi-modal
  integration for building a versatile and reliable molecular assistant in drug
  discovery,'' \emph{arXiv preprint arXiv:2311.16208}, 2023.

\bibitem{liu2023molca}
Z.~Liu, S.~Li, Y.~Luo, H.~Fei, Y.~Cao, K.~Kawaguchi, X.~Wang, and T.-S. Chua,
  ``Molca: Molecular graph-language modeling with cross-modal projector and
  uni-modal adapter,'' \emph{arXiv preprint arXiv:2310.12798}, 2023.

\bibitem{tang2024graphgpt}
J.~Tang, Y.~Yang, W.~Wei, L.~Shi, L.~Su, S.~Cheng, D.~Yin, and C.~Huang,
  ``Graphgpt: Graph instruction tuning for large language models,'' in
  \emph{Proceedings of the 47th International ACM SIGIR Conference on Research
  and Development in Information Retrieval}, 2024, pp. 491--500.

\bibitem{awadalla2024mint}
A.~Awadalla, L.~Xue, O.~Lo, M.~Shu, H.~Lee, E.~K. Guha, M.~Jordan, S.~Shen,
  M.~Awadalla, S.~Savarese \emph{et~al.}, ``Mint-1t: Scaling open-source
  multimodal data by 10x: A multimodal dataset with one trillion tokens,''
  \emph{arXiv preprint arXiv:2406.11271}, 2024.

\bibitem{li2023blip}
J.~Li, D.~Li, S.~Savarese, and S.~Hoi, ``Blip-2: Bootstrapping language-image
  pre-training with frozen image encoders and large language models,'' in
  \emph{International conference on machine learning}.\hskip 1em plus 0.5em
  minus 0.4em\relax PMLR, 2023, pp. 19\,730--19\,742.

\bibitem{li2023characterize}
W.~Li, R.~Purdy, J.~M. Moura, and R.~Blanton, ``Characterize the ability of
  gnns in attacking logic locking,'' in \emph{2023 ACM/IEEE 5th Workshop on
  Machine Learning for CAD (MLCAD)}.\hskip 1em plus 0.5em minus 0.4em\relax
  IEEE, 2023, pp. 1--6.

\bibitem{dubey2024llama}
A.~Dubey, A.~Jauhri, A.~Pandey, A.~Kadian, A.~Al-Dahle, A.~Letman, A.~Mathur,
  A.~Schelten, A.~Yang, A.~Fan \emph{et~al.}, ``The llama 3 herd of models,''
  \emph{arXiv preprint arXiv:2407.21783}, 2024.

\bibitem{devlin2018bert}
J.~Devlin, ``Bert: Pre-training of deep bidirectional transformers for language
  understanding,'' \emph{arXiv preprint arXiv:1810.04805}, 2018.

\bibitem{hu2021lora}
E.~J. Hu, Y.~Shen, P.~Wallis, Z.~Allen-Zhu, Y.~Li, S.~Wang, L.~Wang, and
  W.~Chen, ``Lora: Low-rank adaptation of large language models,'' \emph{arXiv
  preprint arXiv:2106.09685}, 2021.

\bibitem{takamaeda2015pyverilog}
S.~Takamaeda-Yamazaki, ``Pyverilog: A python-based hardware design processing
  toolkit for verilog hdl,'' in \emph{Applied Reconfigurable Computing: 11th
  International Symposium, ARC 2015, Bochum, Germany, April 13-17, 2015,
  Proceedings 11}.\hskip 1em plus 0.5em minus 0.4em\relax Springer, 2015, pp.
  451--460.

\bibitem{hagberg2008exploring}
A.~Hagberg, P.~J. Swart, and D.~A. Schult, ``Exploring network structure,
  dynamics, and function using networkx,'' Los Alamos National Laboratory
  (LANL), Los Alamos, NM (United States), Tech. Rep., 2008.

\bibitem{sweeney2020circuitgraph}
J.~Sweeney, R.~Purdy, R.~D. Blanton, and L.~Pileggi, ``Circuitgraph: A python
  package for boolean circuits,'' \emph{Journal of Open Source Software},
  vol.~5, no.~56, p. 2646, 2020.

\end{thebibliography}
}

\clearpage 
\appendix
\section{Appendix}

\textbf{Appendix A: BRIDGES Dataset generation}

In the following, we cover details of dataset generation, including how each component of a data instance is produced.
\subsubsection*{RTL Code and Function Descriptions}
The workflow begins with RTL-description pairs derived from RTLCoder \cite{liu2024rtlcoder} and MG-Verilog \cite{zhang2024mg}, two open-source datasets for RTL code generation. 
RTLCoder provides 26,532 RTL code instances, which are generated from LLMs prompted by an instruction (step \greencircle{0}). 
We use the instructions to create function descriptions using Llama-3.1-70B \cite{dubey2024llama} with custom prompts (step \greencircle{1}).
MG-Verilog contains 11,144 RTL code instances, each instance is equipped with a multi-grained descriptions, including  simple (high-level) and detailed (block-level) versions. In $\mathsf{BRIDGES}$, detailed descriptions are used for subsequent steps.

\subsubsection*{Dataflow graph}
For each RTL instance, We use PyVerilog \cite{takamaeda2015pyverilog}, a toolkit for processing Verilog HDL, to generate module-level dataflow graphs (step \blackcircle{3}). These graphs, representing different data flows, are further merged and post-processed into NetworkX format \cite{hagberg2008exploring}.

\subsubsection*{Circuit type}
RTLCoder predefines 8 circuit types (see \Cref{fig:histogram_dataset}).
In $\mathsf{BRIDGES}$, we follow the same circuit types, and annotate each data instance with a circuit-type label (step \greencircle{2}).
Specifically, categories are derived from a keyword pool curated by experienced engineers, 
covering common circuit designs. 
Function descriptions are used as prompt to categorize circuits with LLaMA-3-70B and GPT-4o, resulting in two separate circuit-type labels. 

\subsubsection*{Netlist and netlist graph}
We use Genus 22.10 to synthesize RTL designs into Verilog netlists (step \blackcircle{4}) and record the reported area and power.  
We use a simple cell library consisting of two-input logic gates.
Three critical synthesis effort parameters are adjusted to generate diverse results:
1) \texttt{generic\_effort}: Balances quality and runtime by controlling overall synthesis intensity,
2) \texttt{mapping\_effort}: Influences library mapping, affecting timing, area, and power, and
3) \texttt{optimization\_effort}: Controls additional post-mapping QoR (timing or power) optimizations.

Each parameter is configured at three levels (low, medium, high), resulting in 27 unique parameter combinations. 
The synthesis time limit is one hour for each combination. 
 Synthesized netlists are then parsed and converted into directed graphs of primitive gates (step \blackcircle{5}) using CircuitGraph \cite{sweeney2020circuitgraph}, which generates NetworkX graphs.
CircuitGraph may occasionally fail {due to undefined modules in RTL code (undefined modules survived as black boxs through synthesis)}, resulting in fewer than 27 netlist graphs per RTL code.

\noindent \textbf{Appendix B: Prompts}
\subsection{Table I}
\begin{spverbatim}
    Below verilog file is either an adder, comparator, divider, or multiplier, are you able to classify which type it is and what bit it is? Type your answer directly, for example, multiplier-8bit, no analysis.
\end{spverbatim}

\subsection{Dataset generation}

\subsubsection{Step 1}

\begin{spverbatim}
    {"role": "user", "content": """Given a design instruction, change it into a tone of description. Do not change or add any details.  \n
    Here are two examples. \n
    Instruction: Design a module that can detect any edge in an 8-bit binary vector and output the binary value of the vector one cycle after the edge is detected. The module should have two input ports: a clock input and an 8-bit binary input port. The output port should be an 8-bit binary vector that represents the input value one cycle after the edge is detected. The module must be designed using a counter and a comparator.
    \n Example description: This module is designed to detect any edge in an 8-bit binary vector and output the binary value of the vector one cycle after the edge is detected. The module has two input ports: a clock input (clk) and an 8-bit binary input port (in). The output port (out) is an 8-bit binary vector that represents the input value one cycle after the edge is detected. The design uses a counter and a comparator to achieve this functionality.
    \n Instruction: Please act as a professional Verilog designer. Design a pipelined module that implements a 4-to-2 priority encoder. The module should have four 1-bit inputs (I0, I1, I2, I3) and two 2-bit outputs (O0, O1). The output should be the binary encoding of the highest-priority input that is asserted. If multiple inputs are asserted, the output should correspond to the input with the highest index number (i.e., the last asserted input in the list). Use pipeline structure to achieve this functionality.
    \n Example description: This design is a pipelined 4-to-2 priority encoder module. The module has four 1-bit inputs (I0, I1, I2, I3) and two 2-bit outputs (O0, O1). The output is the binary encoding of the highest-priority input that is asserted. If multiple inputs are asserted, the output corresponds to the input with the highest index number. The design uses a pipeline structure to implement this functionality.
    \n Now, please change this instruction directly (do not include any pre-fix like `here is a rewritten description): """ + json_content[i]['instruction']},\end{spverbatim}

\subsubsection{Step 2}

\begin{spverbatim}
    {"role": "system", "content": """
    You are a professional VLSI digital design engineer. Categorize the following RTL (Register Transfer Level) design descriptions and Verilog code pairs into one of the functional categories below. The response should only contain the most relevant function category.
    
    Functional Categories:
    1. Encryption Units: Modules that handle encryption or cryptographic functions.
    2. Data Path Units: Modules involved in data movement, selection, or manipulation (e.g., multiplexers, shifters).
    3. Control Logic Units: Modules responsible for control flow or decision-making in systems (e.g., state machines).
    4. Arithmetic Units: Modules performing arithmetic operations (e.g., adders, subtractors).
    5. Communication Protocol Units: Modules implementing communication protocols (e.g., UART, SPI).
    6. Signal Processing Units: Modules used for signal transformation or filtering.
    7. Clock Management Units: Modules managing clock signals and synchronization.
    8. Other Units: Modules not fitting the above categories.
    Please reply with only the functional category name.
    
    Examples:
    1.
    Description: "This module is a 4-bit adder with carry-in and carry-out. The module has two 4-bit inputs, a single carry-in input, and a single carry-out output. The output is the sum of the two inputs plus the carry-in."
    Verilog: "module adder (\n    input [3:0] a,\n    input [3:0] b,\n    input cin,\n    output cout,\n    output [3:0] sum\n);\n\n    assign {cout, sum} = a + b + cin;\n\nendmodule"
    Response: "Arithmetic Units”
    2.
    Description: "This module is a 2-to-1 multiplexer designed using Verilog. The module has two input ports and one output port. The output is the value of the first input port if the select input is 0, and the value of the second input port if the select input is 1. The design is implemented using only NAND gates."
    Verilog: "module mux_2to1 (\n    input a,\n    input b,\n    input select,\n    output reg out\n);\n\n  wire nand1, nand2, nand3, nand4;\n\n  assign nand1 = ~(a & select);\n  assign nand2 = ~(b & ~select);\n  assign nand3 = ~(nand1 & nand2);\n  assign nand4 = ~(nand3 & nand3);\n\n  always @ (nand4) begin\n    out <= ~nand4;\n  end\n\nendmodule"
    Response: "Data Path Units”
    
    Now categorize the following RTL description and Verilog code pair:
    """
    },
    {"role": "user", "content": f"""
    Description: "{description}"
    Verilog: "{verilog}"
    """}\end{spverbatim}

\subsection{Experiments - type prediction}

\begin{spverbatim}
    Below verilog file is either an adder, comparator, divider, or multiplier, are you able to classify which type it is and what bit it is? Type your answer directly, for example, multiplier-8bit, no analysis.
\end{spverbatim}

\subsubsection{Llama3 - RTL code}
\begin{spverbatim}
    <|begin_of_text|><|start_header_id|>system<|end_header_id|>You are a specialized Verilog code analyzer focused on classifying hardware designs into specific categories. Your task is to analyze Verilog code and determine its primary design type from the following categories:
    Encryption Unit: Designs implementing cryptographic algorithms, secure hash functions, or other security-related operations
    Data Path Unit: Components handling data flow, multiplexers, decoders, registers, and data routing
    Control Logic Unit: State machines, sequence controllers, and decision-making logic
    Arithmetic Unit: Mathematical operations, ALUs, multipliers, dividers, and computational blocks
    Communication Protocol Unit: Implementations of protocols like UART, I2C, SPI, or other communication interfaces
    Signal Processing Unit: Filters, FFT implementations, signal conditioning, and digital signal processing
    Clock Management Unit: Clock generators, PLL implementations, clock dividers, and timing control
    Others: Designs that don't clearly fit into the above categories
    <|eot_id|>
    <|start_header_id|>user<|end_header_id|>
    Please analyze the following Verilog graph and classify it into one of the specified design types. Its RTL code is {}.<|eot_id|>
    <|start_header_id|>assistant<|end_header_id|>
\end{spverbatim}

\subsubsection{Llama3 - netlist code}
\begin{spverbatim}
    <|begin_of_text|><|start_header_id|>system<|end_header_id|>You are a specialized Verilog code analyzer focused on classifying hardware designs into specific categories. Your task is to analyze Verilog code and determine its primary design type from the following categories:
    Encryption Unit: Designs implementing cryptographic algorithms, secure hash functions, or other security-related operations
    Data Path Unit: Components handling data flow, multiplexers, decoders, registers, and data routing
    Control Logic Unit: State machines, sequence controllers, and decision-making logic
    Arithmetic Unit: Mathematical operations, ALUs, multipliers, dividers, and computational blocks
    Communication Protocol Unit: Implementations of protocols like UART, I2C, SPI, or other communication interfaces
    Signal Processing Unit: Filters, FFT implementations, signal conditioning, and digital signal processing
    Clock Management Unit: Clock generators, PLL implementations, clock dividers, and timing control
    Others: Designs that don't clearly fit into the above categories
    <|eot_id|>
    <|start_header_id|>user<|end_header_id|>
    Please analyze the following Verilog graph and classify it into one of the specified design types. Its netlist code is {}.<|eot_id|>
    <|start_header_id|>assistant<|end_header_id|>
\end{spverbatim}

\subsubsection{GPT-4o - RTL code}
\begin{spverbatim}
system_setting = """You are a specialized Verilog code analyzer focused on classifying hardware designs into specific categories.     Your task is to analyze Verilog code and determine its primary design type from the following categories:Encryption Unit: Designs implementing cryptographic algorithms, secure hash functions, or other security-related operationsData Path Unit: Components handling data flow, multiplexers, decoders, registers, and data routingControl Logic Unit: State machines, sequence controllers, and decision-making logicArithmetic Unit: Mathematical operations, ALUs, multipliers, dividers, and computational blocksCommunication Protocol Unit: Implementations of protocols like UART, I2C, SPI, or other communication interfacesSignal Processing Unit: Filters, FFT implementations, signal conditioning, and digital signal processingClock Management Unit: Clock generators, PLL implementations, clock dividers, and timing controlOthers: Designs that don't clearly fit into the above categories""”

messages = [
{"role": "system", "content": system_setting},
{"role": "user", "content": f"""Please analyze the following code and classify it into one of the specified design types. Its RTL code is {data}. Please reply with only the category name. The design type is: """}
]
\end{spverbatim}

\subsubsection{GPT-4o - netlist code}
\begin{spverbatim}
    system_setting = """You are a specialized Verilog code analyzer focused on classifying hardware designs into specific categories.     Your task is to analyze Verilog code and determine its primary design type from the following categories:Encryption Unit: Designs implementing cryptographic algorithms, secure hash functions, or other security-related operationsData Path Unit: Components handling data flow, multiplexers, decoders, registers, and data routingControl Logic Unit: State machines, sequence controllers, and decision-making logicArithmetic Unit: Mathematical operations, ALUs, multipliers, dividers, and computational blocksCommunication Protocol Unit: Implementations of protocols like UART, I2C, SPI, or other communication interfacesSignal Processing Unit: Filters, FFT implementations, signal conditioning, and digital signal processingClock Management Unit: Clock generators, PLL implementations, clock dividers, and timing controlOthers: Designs that don't clearly fit into the above categories""”

    messages = [
    {"role": "system", "content": system_setting},
    {"role": "user", "content": f"""Please analyze the following code and classify it into one of the specified design types. Its netlist code is {data}. Please reply with only the category name. The design type is: """}
    ]
\end{spverbatim}

\subsubsection{$\mathsf{BRIDGES}$}
\begin{spverbatim}
    <|begin_of_text|><|start_header_id|>system<|end_header_id|>
    You are a specialized Verilog code analyzer focused on classifying hardware designs into specific categories. 
    Your task is to analyze Verilog code and determine its primary design type from the following categories:
    Encryption Unit: Designs implementing cryptographic algorithms, secure hash functions, or other security-related operations
    Data Path Unit: Components handling data flow, multiplexers, decoders, registers, and data routing
    Control Logic Unit: State machines, sequence controllers, and decision-making logic
    Arithmetic Unit: Mathematical operations, ALUs, multipliers, dividers, and computational blocks
    Communication Protocol Unit: Implementations of protocols like UART, I2C, SPI, or other communication interfaces
    Signal Processing Unit: Filters, FFT implementations, signal conditioning, and digital signal processing
    Clock Management Unit: Clock generators, PLL implementations, clock dividers, and timing control
    Others: Designs that don't clearly fit into the above categories
    <|eot_id|>
    <|start_header_id|>user<|end_header_id|>
    Please analyze the following Verilog graph and classify it into one of the specified design types. Its graph tokens are {}.<|eot_id|>
    <|start_header_id|>assistant<|end_header_id|>
\end{spverbatim}

\subsection{Experiments - function description}

\subsubsection{Llama3 - RTL code}
\begin{spverbatim}
    <|begin_of_text|><|start_header_id|>system<|end_header_id|>You are a hardware description expert. Provide a single, coherent technical paragraph describing the functionality of a Verilog module.
    Constraints:
    - Use complete English sentences.
    - Avoid mentioning variable names or including any Verilog syntax.
    - Ensure the description focuses on functionality, not implementation details.
    - Do not use lists, bullet points, or code snippets.
    - Maintain a logical flow without line breaks or special formatting.
    
    Example:
    ---
    Module Description:
    This module implements an edge detection mechanism. It accepts an 8-bit binary input and a clock signal,
    producing an 8-bit output that reflects the input value one cycle after an edge is detected.
    The circuit operates by comparing the current input with the previous input to identify edges, utilizing a counter to manage the delay in output generation.
    ---
    <|eot_id|>
    <|start_header_id|>user<|end_header_id|>
    Provide a detailed description of the following Verilog module. Its RTL code is {}. <|eot_id|>
    <|start_header_id|>assistant<|end_header_id|>
\end{spverbatim}

\subsubsection{Llama3 - netlist code}
\begin{spverbatim}
    <|begin_of_text|><|start_header_id|>system<|end_header_id|>You are a hardware description expert. Provide a single, coherent technical paragraph describing the functionality of a Verilog module.
    Constraints:
    - Use complete English sentences.
    - Avoid mentioning variable names or including any Verilog syntax.
    - Ensure the description focuses on functionality, not implementation details.
    - Do not use lists, bullet points, or code snippets.
    - Maintain a logical flow without line breaks or special formatting.
    
    Example:
    ---
    Module Description:
    This module implements an edge detection mechanism. It accepts an 8-bit binary input and a clock signal,
    producing an 8-bit output that reflects the input value one cycle after an edge is detected.
    The circuit operates by comparing the current input with the previous input to identify edges, utilizing a counter to manage the delay in output generation.
    ---
    <|eot_id|>
    <|start_header_id|>user<|end_header_id|>
    Provide a detailed description of the following Verilog module. Its verilog code is {}. <|eot_id|>
    <|start_header_id|>assistant<|end_header_id|>
\end{spverbatim}

\subsubsection{$\mathsf{BRIDGES}$ w. RTL code}
\begin{spverbatim}
    <|begin_of_text|><|start_header_id|>system<|end_header_id|>You are a hardware description expert. Provide a single, coherent technical paragraph describing the functionality of a Verilog module.
    Constraints:
    - Use complete English sentences.
    - Avoid mentioning variable names or including any Verilog syntax.
    - Ensure the description focuses on functionality, not implementation details.
    - Do not use lists, bullet points, or code snippets.
    - Maintain a logical flow without line breaks or special formatting.
    
    Example:
    ---
    Module Description:
    This module implements an edge detection mechanism. It accepts an 8-bit binary input and a clock signal,
    producing an 8-bit output that reflects the input value one cycle after an edge is detected.
    The circuit operates by comparing the current input with the previous input to identify edges, utilizing a counter to manage the delay in output generation.
    ---
    <|eot_id|>
    <|start_header_id|>user<|end_header_id|>
    Provide a detailed description of the following Verilog module. Its RTL code is {}. Its graph representations are {}.<|eot_id|>
    <|start_header_id|>assistant<|end_header_id|>
\end{spverbatim}

\subsubsection{$\mathsf{BRIDGES}$}
\begin{spverbatim}
    <|begin_of_text|><|start_header_id|>system<|end_header_id|>You are a hardware description expert. Provide a single, coherent technical paragraph describing the functionality of a Verilog module.
    Constraints:
    - Use complete English sentences.
    - Avoid mentioning variable names or including any Verilog syntax.
    - Ensure the description focuses on functionality, not implementation details.
    - Do not use lists, bullet points, or code snippets.
    - Maintain a logical flow without line breaks or special formatting.
    
    Example:
    ---
    Module Description:
    This module implements an edge detection mechanism. It accepts an 8-bit binary input and a clock signal,
    producing an 8-bit output that reflects the input value one cycle after an edge is detected.
    The circuit operates by comparing the current input with the previous input to identify edges, utilizing a counter to manage the delay in output generation.
    ---
    <|eot_id|>
    <|start_header_id|>user<|end_header_id|>
    Provide a detailed description of the following Verilog module. Its graph tokens are {}.<|eot_id|>
    <|start_header_id|>assistant<|end_header_id|>
\end{spverbatim}

\end{document}